\documentclass[sigconf, table]{acmart}

\usepackage{booktabs} 
\usepackage{xcolor} 
\usepackage{eso-pic}
\usepackage{xspace}
\usepackage{cleveref}
\usepackage{times}
\usepackage{float}
\usepackage{tcolorbox}
\usepackage{epsfig}
\usepackage{graphicx}
\usepackage{amsmath}
\usepackage{amssymb}
\usepackage{color}
\usepackage{url}
\usepackage{tabu}
\usepackage{multirow}
\usepackage{pifont}
\usepackage{tikz}
\usepackage{pgfplots}
\usepackage{tikz}
\usepgfplotslibrary{external}
\usetikzlibrary{positioning}
\graphicspath{{./images/}}

\newcommand*{\ityping}{\includegraphics[scale=0.02]{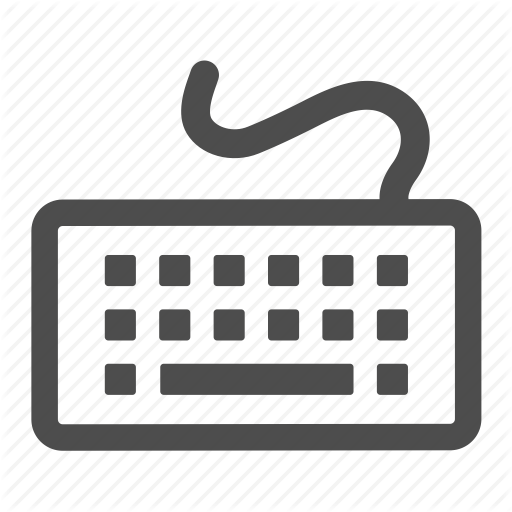}}
\newcommand*{\ispeaking}{\includegraphics[scale=0.03]{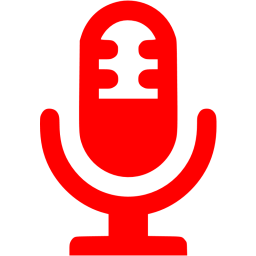}}

\begin{document}
\title{Speech-Based Visual Question Answering}

\author{Ted Zhang}
\affiliation{%
  \institution{KU Leuven}
}
\email{tedz.cs@gmail.com}

\author{Dengxin Dai}
\affiliation{%
  \institution{ETH Zurich}
}
\email{dai@vision.ee.ethz.ch}

\author{Tinne Tuytelaars}
\affiliation{%
  \institution{KU Leuven}
}
\email{tinne.tuytelaars@esat.kuleuven.be }

\author{Marie-Francine Moens}
\affiliation{%
  \institution{KU Leuven}
}
\email{sien.moens@cs.kuleuven.be}

\author{Luc Van Gool}
\affiliation{%
  \institution{ETH Zurich, KU Leuven}
}
\email{vangool@vision.ee.ethz.ch}

\begin{abstract}
This paper introduces speech-based visual question answering (VQA), the task of generating an answer given an image and a spoken question. Two methods are studied: an end-to-end, deep neural network that directly uses audio waveforms as input versus a pipelined approach that performs ASR (Automatic Speech Recognition) on the question, followed by text-based visual question answering. Furthermore, we investigate the robustness of both methods by injecting various levels of noise into the spoken question and find both methods to be tolerate noise at similar levels.

\end{abstract}
%
%
%


\maketitle
\label{sec:intro}
\section{Introduction}
The recent years have witnessed great advances in computer vision, natural language processing, and speech recognition thanks to the advances in deep learning \cite{lecun2015deep} and abundance of data \cite{imagenet:2015}. This is evidenced not only by the surge of academic papers, but also by the world-wide industry interests. The convincing successes in these individual fields naturally raise the potentials of further integration towards solutions to more general AI problems. Much work has been done to integrate vision and language, resulting in a wide collection of successful applications such as image/video captioning \cite{show:tell:caption}, movie-to-book alignment \cite{align:bookmovie}, and visual question answering (VQA) \cite{VQA}. However, the importance of integrating vision and speech has remained relatively unexplored.

Pertaining to practical applications, voice-user interface (VUI) has become more commonplace, and people are increasingly taking advantage of its characteristics; it is natural, hands-free, eyes-free, far more mobile and faster than typing on certain devices \cite{speech:faster}. As many of our daily tasks are relevant to visual scenes, there is a strong need to have a VUI to talk to pictures or videos directly, be it for communication, cooperation, or guidance. Speech-based VQA can be used to assist blind people in performing ordinary tasks, and to dictate robotics in real visual scenes in a hand-free manner such as clinical robotic surgery.

\begin{figure}[t]
\centering
\begin{tikzpicture}[
bluesquare/.style={rectangle, draw=blue!70, fill=blue!4, very thick, minimum size=5mm},
greensquare/.style={rectangle, draw=green!70, fill=green!4, very thick, minimum size=5mm},
magsquare/.style={rectangle, draw=magenta!70, fill=magenta!4, very thick, minimum size=5mm},
empty/.style={rectangle, very thick, minimum size=7mm},
]

\node[magsquare, outer sep=5pt]      (model2)           {TextMod};
\node[empty] (answer2) [right=of model2] {pizza};
\node[inner sep=5pt] (pizza2)  [above=of model2] {\includegraphics[width=.1\textwidth]{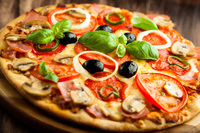}};
\node[greensquare, outer sep=5pt, label=below:what food is this?]    (asr)  [left=1cm of model2] {ASR}; 
\node[inner sep=5pt] (wav2)  [above=of asr] {\includegraphics[width=.1\textwidth]{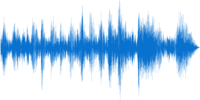}}; 

\node[inner sep=5pt] (pizza1)  [below=of model2] {\includegraphics[width=.1\textwidth]{pizzas.jpg}};
\node[bluesquare, outer sep=5pt]   [below=of pizza1]   (model1)                      {SpeechMod};
\node[empty] (answer1) [right=of model1] {pizza};
\node[inner sep=5pt] (wav1)  [left=of model1] {\includegraphics[width=.1\textwidth]{wavs.png}};

\draw[->] (wav1.east) -- (model1.west);
\draw[->] (pizza1.south) -- (model1.north);
\draw[->] (model1.east) -- (answer1.west);

\draw[->] (wav2.south) -- (asr.north);
\draw[->] (asr.east) -- (model2.west);
\draw[->] (pizza2.south) -- (model2.north);
\draw[->] (model2.east) -- (answer2.west);

\end{tikzpicture}
\caption{An example of speech-based visual question answering and the two method in this study. A spoken question \emph{what food is this?} is asked about the picture, and the system is expected to generate the answer \emph{pizza}.}
\label{fig:models}
\end{figure}

This work investigates the potential of integrating vision and speech in the context of VQA. A spoken version of the \textit{VQA1.0} dataset is generated to study two different methods of speech-based question answering. One method is an end-to-end approach based on a deep neural network architecture, and the other uses an ASR to first transcribe the text from the spoken question, as shown in \Cref{fig:models}. The former approach can be particularly useful for languages that are not serviced by popular ASR systems, i.e. minor languages that have scarce text-speech aligned training data.

The \textbf{main contributions} of this paper are three-fold: \textbf{1)} We introduce an end-to-end model that directly produces answers from auditory input without transformations into intermediate pre-learned representations, and compare this with the pipelined approach. \textbf{2)} We inspect the performance impact of having different levels of background noise mixed with the original utterances. \textbf{3)} We release the speech dataset, roughly 200 hours of synthetic audio data and 1 hour of real speech data, to the public.~\footnote{http://data.vision.ee.ethz.ch/daid/VQA/SpeechVQA.zip} The emphasis of this paper is not on achieving state of the art numbers on VQA, but rather on exploring ways to address a new and challenging task.

\section{Related Works}

\subsection{Visual Question Answering}
The initial introduction of VQA into the AI community \cite{realtime:vqa}, \cite{daquar} was motivated by a desire to build intelligent systems that can understand the world more holistically. In order to complete the task of VQA, it was both necessary to understand a textual question and a visual scene. However, it was not until the introduction of \textit{VQA1.0} \cite{VQA} that the application took mainstream in the computer vision and natural language processing (NLP) communities.

Recently, popular topics of exploration have been on the development of attention models. Attention models were popularized by their success with the NLP community in machine translation \cite{nmt:joint:align}, and quickly demonstrated their efficacy in computer vision \cite{recurrent:visual:attn}. Within the context of visual question answering, attention mechanisms `show' a model where to look when answering a question. Stacked Attention Network \cite{vqa:stackedattn} learns an attention mechanism based on the of the question's encoding to determine the salient regions in an image. More sophisticated attention-centric models such as \cite{vqa:dualattn,vqa:hieco,vqa:dynamicmemory} were since then developed.

Other points of research are based on the pooling mechanism that combines the language component with the vision components. Some use an element-wise multiplication \cite{vqa:spatial:attn,vqa:stackedattn} to pool these modalities, while \cite{vqa:hieco} and \cite{multimodal:pooling} have shown much success in using more complex methods. Our work differs from theirs in that we aim not to improve the performance of VQA, but rather add a new modality of input and introduce appropriate new methods.

\subsection{Integration of Speech and Vision}
The works also relevant to ours are those integrating speech and vision. Pixeltone \cite{pixel:tone}
and Image spirit \cite{image:spirit} are examples that use voice
commands to guide image processing and semantic segmentation. There is also academic work \cite{show:tell,speech:anno:img,speech:retri:img} and an app \cite{smile} that use speech to provide image descriptions. Their tasks and algorithms are both different from ours. We study the potential of integrating speech and vision in the context of VQA and aim to learn a joint understanding of speech and vision. Those approaches, however, use speech recognition for data collection or result refinement.

Our work also shares similarity with visual-grounded speech understanding or recognition. The closest one in this vein is \cite{speech:caption}, in which a deep model is learned with speeches about image captions for speech-based image retrieval. In a broader context of integration of sound and vision, Soundnet \cite{soundnet} transfers visual information into sound representations, but this differs from our work because their end goal is to label a sound, not to answer a question.

\subsection{End-To-End Speech Recognition}
In the past decade, deep learning has allowed many fields in artificial intelligence to replace traditional hand-crafted features and pipeline systems with end-to-end models. Since speech recognition is typically thought of as a sequence to sequence transduction problem, i.e. given an input sequence, predict an output sequence, the application of LSTM and the CTC \cite{rnn:discrimspotting,ctc} promptly showed the success needed to justify its superiority over traditional methods. Current state of the art ASR systems such as DeepSpeech2 \cite{deepspeech2} uses stacked Bi-directional Recurrent Neural Networks in conjunction with Convolutional Neural networks. Our model is similar to theirs in that we use CNNs connected with an LSTM to process audio inputs, however our goal is question answering and not speech recognition.

\begin{table*}[!t]
\centering
\caption{Dimensions for the conv layers. Example shown with a 2 second long audio waveform, sampled at 16 kHz. The final output dimensions are (3, 512)}
\label{table:convdims}
\begin{tabular}{l|c|c|c|c|c|c|c|c|c}
Layer         & \multicolumn{1}{l|}{conv1} & \multicolumn{1}{l|}{pool1} & \multicolumn{1}{l|}{conv2} & \multicolumn{1}{l|}{pool2} & \multicolumn{1}{l|}{conv3} & \multicolumn{1}{l|}{pool3} & \multicolumn{1}{l|}{conv4} & \multicolumn{1}{l|}{pool4} & \multicolumn{1}{l}{conv5} \\ \hline
Input Dim     & 32,000                      & 16,000                      & 4,000                       & 2,000                       & 500                        & 250                        & 62                         & 31                         & 7                         \\
\# Filters    & 32                         & 32                         & 64                         & 64                         & 128                        & 128                        & 256                        & 256                        & 512                       \\
Filter Length & 64                         & 4                          & 32                         & 4                          & 16                         & 4                          & 8                          & 4                          & 4                         \\
Stride        & 2                          & 4                          & 2                          & 4                          & 2                          & 4                          & 2                          & 4                          & 2                         \\
Output Dim    & 16,000                      & 4,000                       & 2,000                       & 500                        & 250                        & 62                         & 31                         & 7                          & 3                        
\end{tabular}
\end{table*}

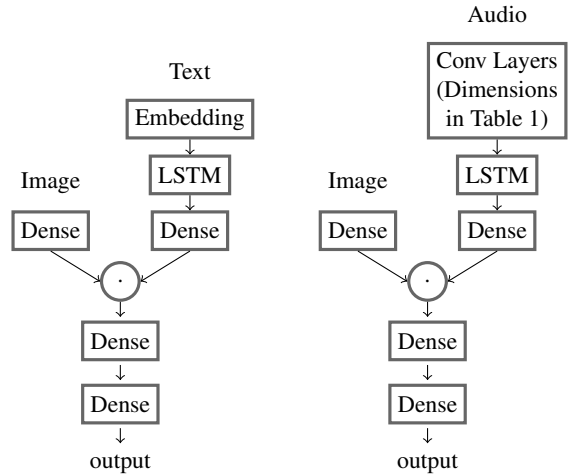
\begin{figure}[!t]
\centering
\begin{tikzpicture}[
inps/.style={rectangle, draw=black!60, very thick, minimum size=10mm},
rect/.style={rectangle, draw=black!60, very thick, minimum size=5mm},
circ/.style={circle, draw=black!60, very thick, minimum size=5mm},
empty/.style={rectangle, very thick, minimum size=5mm},
]

\node[empty]      (central)             {};

\node[circ]      (1main)          [left =15mm of central]      {.};
\node[rect]      (1ldense)        [above right=3mm of 1main]        {Dense};
\node[rect]      (1lstm)     [above=2mm of 1ldense]           {LSTM};
\node[rect]      (1emb)          [above=2mm of 1lstm]      {Embedding};
\node[empty]      (1linput)    [above=1mm of 1emb]            {Text};
\node[rect]      (1vdense)       [above left=3mm of 1main]         {Dense};
\node[empty]        (1vinput)  [above=1mm of 1vdense] {Image};
\node[rect, outer sep=2pt]      (1cdense1)        [below=2mm of 1main]        {Dense};
\node[rect, outer sep=2pt]      (1cdense2)        [below=2mm of 1cdense1]        {Dense};
\node[empty] (1answer) [below=2mm of 1cdense2] {output};

\draw[->] (1emb.south) -- (1lstm.north);
\draw[->] (1lstm.south) -- (1ldense.north);
\draw[->] (1ldense.south) -- (1main.east);

\draw[->] (1vdense.south) -- (1main.west);

\draw[->] (1main.south) -- (1cdense1.north);
\draw[->] (1cdense1.south) -- (1cdense2.north);
\draw[->] (1cdense2.south) -- (1answer.north);

\node[circ]      (2main)          [right =15mm of central]      {.};
\node[rect]      (2ldense)        [above right=3mm of 2main]        {Dense};
\node[rect]      (2lstm)     [above=2mm of 2ldense]           {LSTM};
\node[rect]      (convs)          [above=2mm of 2lstm, align=center]      {Conv Layers\\(Dimensions\\in \Cref{table:convdims})};
\node[empty]    (2linput)         [above=1mm of convs]      {Audio};
\node[rect]      (2vdense)       [above left=3mm of 2main]         {Dense};
\node[empty]     (2vinput)  [above=1mm of 2vdense] {Image};
\node[rect, outer sep=2pt]      (2cdense1)        [below=2mm of 2main]        {Dense};
\node[rect, outer sep=2pt]      (2cdense2)        [below=2mm of 2cdense1]        {Dense};
\node[empty] (2answer) [below=2mm of 2cdense2] {output};

\draw[->] (convs.south) -- (2lstm.north);
\draw[->] (2lstm.south) -- (2ldense.north);
\draw[->] (2ldense.south) -- (2main.east);

\draw[->] (2vdense.south) -- (2main.west);

\draw[->] (2main.south) -- (2cdense1.north);
\draw[->] (2cdense1.south) -- (2cdense2.north);
\draw[->] (2cdense2.south) -- (2answer.north);
\end{tikzpicture}
\caption{TextMod (left) and SpeechMod (right) architectures}
\label{fig:modarc}
\end{figure}

\section{Model}
Two models are employed in this work, they will be referred to henceforth as TextMod and SpeechMod. TextMod and SpeechMod only differ in their language components, keeping rest of the architecture the same. On the language side, TextMod takes as input a series of one-hot encodings, followed by an embedding layer that is learned from scratch, a LSTM encoder, and a dense layer. It is similar to \textit{VQA1.0} with some minor adjustments.

The language side of SpeechMod takes as input the raw waveform, and pushes it through a series of 1D convolutions. After the CNN layers follows a LSTM. The LSTM serves the same purpose as in TextMod, which is to interpret and encode the sequence meaningfully into a single vector.

Convolution layers are used to encode waveforms because they reduce dimensionality of data while finding salient patterns. The maximum length of a spoken question in our dataset is 6.7 seconds and corresponds to a waveform length of 107,360 elements, while the minimum is 0.63 seconds and corresponds to 10,080 elements. One could directly feed the input waveform to a LSTM, but a LSTM will be unable to learn from sequences that are excessively long, so dimensionality reduction is a necessity. Each consecutive convolution layer halves in filter length but doubles the number of filters. This is done for simplicity rather than for performance optimization. The main consideration taken in choosing the parameters is that the last convolution should output dimensions of ($x$, 512), where $x$ must be a positive integer. $x$ represents the length of a sequence of 512-dim vectors. The sequence is then fed into an LSTM, which then outputs a single vector of (512). Thus, $x$ should not be too big, and the CNN parameters are chosen to ensure a sensible sequence length. The exact dimensions of the convolution layers are shown in \Cref{table:convdims}. The longest and shortest waveforms correspond to final convolution outputs of size (13, 512) and (1, 512) respectively. 512 is used as the dimension of the LSTM to be consistent with TextMod and the original VQA baseline.

On the visual side, both models ingest as input the 4,096 dimensional vector of the last layer of VGG19 \cite{vgg16} followed by a single dense layer. After both visual and linguistic representations are computed, they are merged using element-wise multiplication, a dense layer, and an output layer. The full architecture of both these models are seen in \Cref{fig:modarc}, where $\displaystyle\odot$ is the symbol for element-wise multiplication. After merging the language and visual components of each model, two dense layers are stacked. The last dense layer outputs a probability distribution over the number of output classes, and the answer corresponding to the element with the highest probability is selected.

The architectures presented in this chapter were chosen for two main reasons: simplicity and similarity. First, the intention is to keep the model complexity low. In order to establish a baseline for speech-based VQA, it is necessary to use only the bare minimum components. TextMod, as mentioned before, is similar to the original VQA baseline, which is well referenced and remains the simplest architecture on \textit{VQA1.0}. Despite its many convolution layers, SpeechMod also uses minimal components. Second, it is important that TextMod and SpeechMod differ from each other as little as possible. Similarity between models allows one to locate the source of discrepancies and helps produce a more rigorous comparison. The only difference in the two models is replacing an embedding layer with a series of convolution layers. In our implementation, the layers that are common between the two models also have the same dimensions.

\section{Data}
We chose to use \textit{VQA1.0} Open-Ended dataset, for its numerous training examples and familiarity to those working in question answering. To avoid confusion, \textit{VQA1.0} henceforth refers to the dataset and the original paper, while VQA refers to the task of visual question answering. The dataset contains 248,349 questions in the training set, 121,512 in validation set, and 60,864 in the test-dev set. The complete test set contains 244,302 questions, but because the evaluation server allows for only one submission, we instead evaluate on test-dev, which has no such limit. During training, questions which do not contain the 1000 most common answers are filtered out.

Amazon Polly API is used to generate audio files for each question.~\footnote{The voice of Joanna is used: \url{https://aws.amazon.com/polly/}} The generated speech is in mp3 format, then sampled into waveform format at 16 kHz. 16 kHz was chosen due to its common usage among the speech community, but also because the model used to transcribe speech was trained on 16 kHz audio waveforms. It is worthwhile to note that the audio generator uses a female voice, thus the training and testing data are all with the same voice, except for the examples we’ve recorded, which is covered below. The full Amazon Polly speech dataset will be made available to the public.

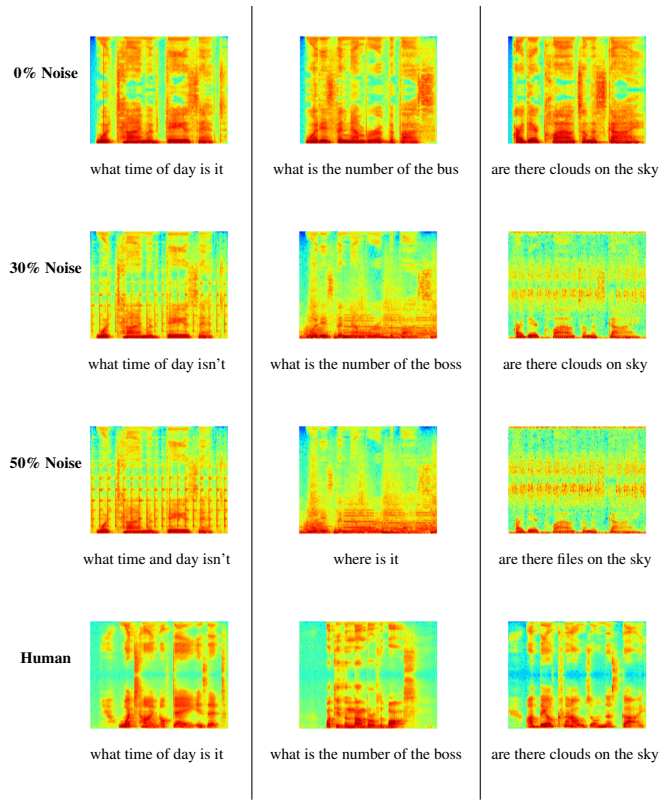
\begin{figure}[t]
\centering
\begin{tikzpicture}[
scale=0.66,
every node/.style={scale=0.66},
squarednode/.style={rectangle, draw=red!70, fill=red!4, very thick, minimum size=5mm},
empty/.style={rectangle, very thick, minimum size=1mm},
]
\node (-center) [empty]      (model)                      {};

\node[inner sep=0.5pt, label=below:what is the number of the bus] (2n0)  [below=8mm of model] {\includegraphics[width=0.2\textwidth]{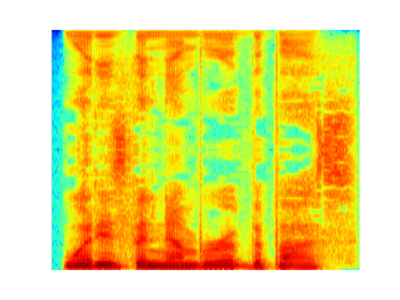}};
\node[inner sep=0.5pt, label=below:what is the number of the boss] (2n30)  [below=8mm of 2n0] {\includegraphics[width=0.2\textwidth]{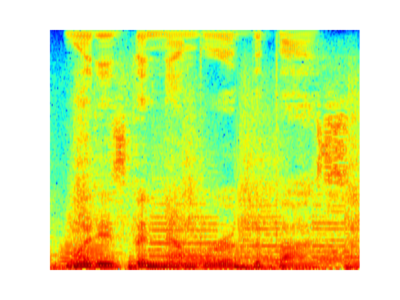}};
\node[inner sep=0.5pt, label=below:where is it] (2n50)  [below=8mm of 2n30] {\includegraphics[width=0.2\textwidth]{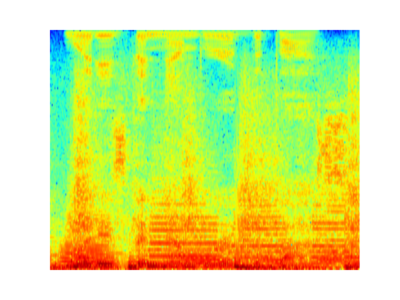}};
\node[inner sep=0.5pt, label=below:what is the number of the boss] (2r)  [below=8mm of 2n50] {\includegraphics[width=0.2\textwidth]{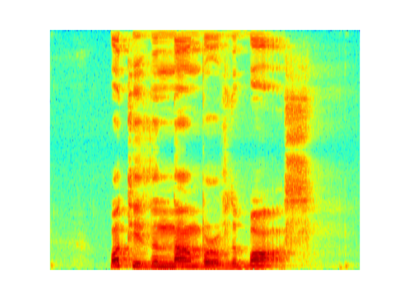}};

\node[inner sep=0.5pt, label=below:what time of day is it] (1n0)  [left=4mm of 2n0] {\includegraphics[width=0.2\textwidth]{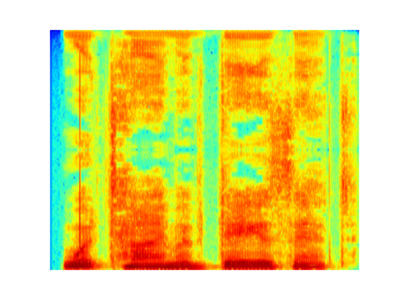}};
\node[inner sep=0.5pt, label=below:what time of day isn't] (1n30)  [left=4mm of 2n30] {\includegraphics[width=0.2\textwidth]{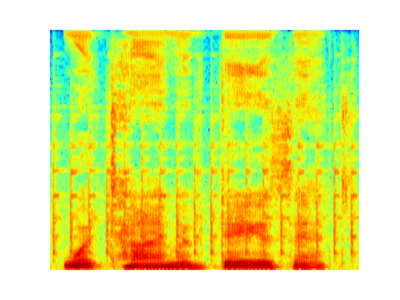}};
\node[inner sep=0.5pt, label=below:what time and day isn't] (1n50)  [left=4mm of 2n50] {\includegraphics[width=0.2\textwidth]{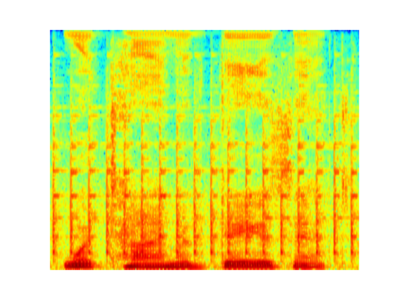}};
\node[inner sep=0.5pt, label=below:what time of day is it] (1r)  [left=4mm of 2r] {\includegraphics[width=0.2\textwidth]{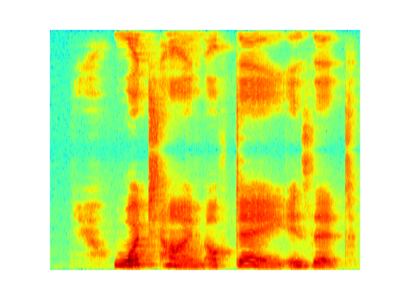}};

\node[inner sep=0.5pt, label=below:are there clouds on the sky] (3n0)  [right=4mm of 2n0] {\includegraphics[width=0.2\textwidth]{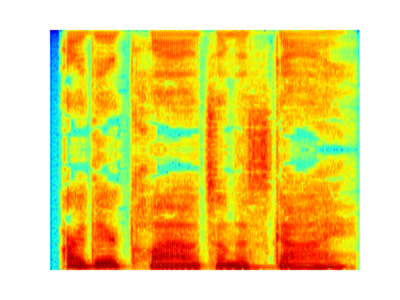}};
\node[inner sep=0.5pt, label=below:are there clouds on sky] (3n30)  [right=4mm of 2n30] {\includegraphics[width=0.2\textwidth]{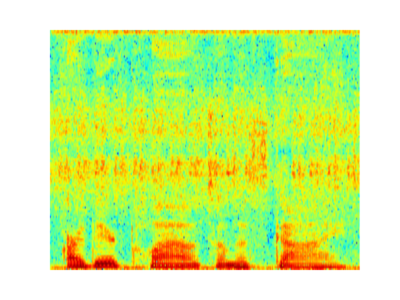}};
\node[inner sep=0.5pt, label=below:are there files on the sky] (3n50)  [right=4mm of 2n50] {\includegraphics[width=0.2\textwidth]{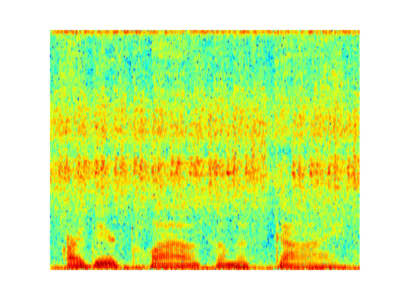}};
\node[inner sep=0.5pt, label=below:are there clouds on the sky] (3r)  [right=4mm of 2r] {\includegraphics[width=0.2\textwidth]{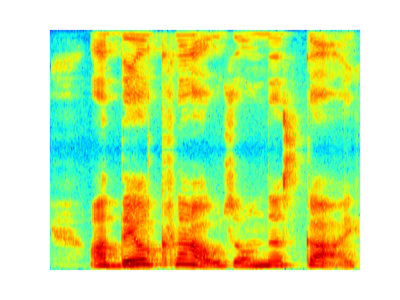}};

\node (-center) [empty, label=\textbf{0\% Noise}]  (n0label)  [left=2mm of 1n0]     {};
\node (-center) [empty, label=\textbf{30\% Noise}] (n30label) [left=2mm of 1n30]    {};
\node (-center) [empty, label=\textbf{50\% Noise}] (n50label) [left=2mm of 1n50]    {};
\node (-center) [empty, label=\textbf{Human}]      (hmlabel)  [left=2mm of 1r]      {};

\draw (-2.3,-1) -- (-2.3,-17);
\draw (2.3,-1) -- (2.3,-17);
\end{tikzpicture}
\caption{Spectrograms for 3 example questions with corresponding transcribed text below. 3 synthetically generated and 1 human-recorded audio clips for each question.}
\label{fig:specs}
\end{figure}

The noise we mixed with the original speech files is selected randomly from the Urban8K dataset \cite{urban8k}. This dataset contains 10 categories: air conditioner, car horn, children playing, dog bark, drilling, engine idling, gun shot, jackhammer, siren, and street music. Some clips are soft enough in volume and thus considered background noise, others are loud enough to be considered foreground noise. For each original audio file, a random noise file is selected, and combined to produce a corrupted question file according to the weighting scheme:
\begin{displaymath}
  W_{corrupted} = (1-\mathit{NL})*W_{original}  + \mathit{NL}*W_{noise}
\end{displaymath}
where \textit{NL} is the noise level. The noise audio files are subsampled to 16 kHz in order to match that of the original audio file, and is clipped to also match the spoken question length. When the spoken question is longer than the noise file, the noise file is repeated until its duration exceeds that of the spoken question. Both files are normalized before being combined so that contributions are strictly proportional to the noise level chosen. We choose 5 noise levels to mix together: 10\%-50\%, at 10\% intervals. Anything beyond 50\% is unrealistic. A visualization of different noise levels can be seen in \Cref{fig:specs} and its corresponding audio clips can be found online.\footnote{https://soundcloud.com/sbvqa/sets/speechvqa}

We also make an additional, supplementary study of the practicality of speech-based VQA with real data. 1000 questions from the \textit{val} set were randomly selected and recorded with human speakers. Two speakers (one male and one female) participated the recording task. In total, 1/3 of the data is from a male speaker, the rest is from a female speaker. Both speakers are graduate students who are not native anglophones. The data was recorded in an office environment, and there are various background noises in the audio clips as they naturally occurred.

\begin{table}[t]
\centering
\caption{Word Error Rate from Kaldi speech recognition}
\label{table:wer}
\begin{tabular}{c|c}
Noise (\%)   & WER (\%) \\ \hline
0  & 8.46  \\
10 & 12.37 \\
20 & 17.77 \\
30 & 25.41 \\
40 & 35.15 \\
50 & 47.90
\end{tabular}
\end{table}

\section{Experiments}
\subsection{Preprocessing}
For SpeechMod, the first preprocessing step is to scale each waveform to a range of [-256, 256], similar to the procedure from SoundNet \cite{soundnet}. There was no need to center each example around 0, as they are already centered. Next, each batch of waveforms were padded with 0 at the end to be of the same length.

For TextMod, the standard preprocessing steps from \textit{VQA1.0} were followed. The procedure tokenizes each sentence and replaces it with a number that corresponds to the word's index. These number indices are used as input, since the question will be fed to the model as a sequence of one hot encodings. Because questions have different lengths, the 0 index is used as padding for sequences that are too short. The 0 index essentially causes the model to skip that position. 0 is also used for unseen tokens, which is especially useful when dealing with out of vocabulary words during evaluation.

\subsection{ASR}
We use Kaldi \cite{kaldi} for ASR, due to its open-source codebase and popularity with the speech research community. The model used in this work is a DNN-HMM\footnote{https://github.com/api-ai/api-ai-english-asr-model} that has been pre-trained on assistant.ai logs (essentially short commands), making it suitable for transcribing short utterances such as the questions in \textit{VQA1.0}. Other ASRs such as wit.ai from Facebook, Cloud Speech from Google, and Bing Speech Microsoft were tested but not used in the final experiments because Kaldi achieved the lowest word error rates.

Word error rate (WER) is used to measure the accuracy of speech to text transcriptions. WER is defined as follows: 
\begin{displaymath}
  \mathit{WER} = (S+D+I)/N
\end{displaymath}
Where \textit{S} is the number of substitutions, \textit{D} is the number of deletions, and \textit{I} is the number of insertions. \textit{N} is the total number of words in the sentence being translated. Each transcribed question is compared with the original; the results are shown in \Cref{table:wer}. WER is not expected to be a perfect measure of transcription accuracy, since some words are more essential to the meaning of a sentence than other words. For example, missing the word \emph{dog} in the sentence \emph{what is the dog eating} is more detrimental than missing the word \emph{the}, but we nevertheless employ it to convey a general notion of how many words are understood by the ASR. Naturally the more noise there is, the higher the word error rate becomes. Due to transcription errors, there are resulting questions that contain words not seen in the original datasets. These words, as mentioned above, are indexed as 0 and are masked when fed into TextMod.

\subsection{Implementation}
Keras was used to run all experiments, with the Adam \cite{kingma2014adam} optimizer for both architectures. No parameter tuning was done; default Adam parameters are as follows: learning rate=0.001, beta1=0.9, beta2=0.999, epsilon=1e-08, learning  rate decay=0.0. Training TextMod for 10 epochs on \textit{train} + \textit{val} takes roughly an hour on a Nvidia Titan X GPU, and our best model was taken at 30 epochs. Training SpeechMod for 10 epochs takes roughly 7 hours. The reported model is taken at 30 epochs. The code is available to the public.\footnote{https://github.com/zted/sbvqa}

\section{Results}
The goal of the main experiments were to observe how each model performs with and without different levels of noise added. Results are reported on \textit{test-dev}, which corresponds to training on \textit{train} + \textit{val} (\Cref{table:vqa test-dev}). The standard format of reporting results from \textit{VQA1.0} is followed: \textit{All} is the overall accuracy, \textit{Y/N} is for questions with yes or no as answers, \textit{Number} is for questions that are answered by counting, and \textit{Other} covers the rest.

TextMod is trained on the original questions (\textit{OQ}), with the best performing model being selected. ASR is used on the 0-50\% variants to convert the audio question to text. Then, the selected model from \textit{OQ} is used to evaluate based on the transcribed text. Concretely, the best performing model obtained on \textit{test-dev} is used to evaluate the transcribed variants of \textit{test-dev}. Likewise, SpeechMod is first trained on audio data with 0\% noise, with the strongest model being selected. The selected model is used to evaluate on the 10-50\% variants of the same data subset. Typically, the best model on \textit{val} is used to evaluate on \textit{test} or another `unseen' portion of the dataset. However in these experiments, the noisy variants of the same datasets are in fact unseen because the data for which the model is trained on contains no noise. We show this in the zero-shot section of the paper.

\begin{table}[t]
\centering
\caption{Accuracy on \textit{test-dev} with different levels of noise added. (Higher is better)}
\label{table:vqa test-dev}
\begin{tabular}{ll|cccc}
          &                     & All    & Y/N    & Number & Other \\ \hline
Baseline  &                     & 53.74  & 78.94  & 35.24  & 36.42 \\ \hline
TextMod   & Blind               & 48.76  & 78.20  & 35.68  & 26.59 \\
          & OQ                  & 56.66  & 78,89  & 37.24  & 42.07 \\
          & 0\%                 & 54.03  & 75.47  & 36.82  & 39.62 \\
          & 10\%                & 52.56  & 74.06  & 36.50  & 37.85 \\
          & 20\%                & 50.22  & 71.16  & 35.72  & 35.64 \\
          & 30\%                & 47.03  & 67.31  & 34.45  & 32.56 \\
          & 40\%                & 42.83  & 62.35  & 31.97  & 28.64 \\
          & 50\%                & 37.12  & 25.42  & 27.05  & 23.77 \\ \hline
SpeechMod & Blind               & 42.05  & 70.85  & 31.62  & 19.84 \\ 
          & 0\%                 & 46.99  & 67.87  & 30.84  & 32.82 \\
          & 10\%                & 45.81  & 67.29  & 30.13  & 31.03 \\
          & 20\%                & 43.33  & 65.88  & 29.24  & 27.28 \\
          & 30\%                & 40.07  & 64.15  & 27.82  & 22.28 \\
          & 40\%                & 35.85  & 61.47  & 24.68  & 16.52 \\
          & 50\%                & 32.14  & 59.33  & 20.84  & 11.50 
\end{tabular}
\end{table}

\textit{Blind} denotes no visual information, meaning it removes the visual components while rest of the model stays the same. TextMod \textit{Blind} is trained and evaluated on the original questions. SpeechMod \textit{Blind} is trained and evaluated on the 0\% noise audio. \textit{Baseline} is from \textit{VQA1.0} using the model `LSTM Q+I'.

\begin{figure*}[t]
$\begin{tabular}{cccc}
\includegraphics[width=0.24\linewidth, height=30mm]{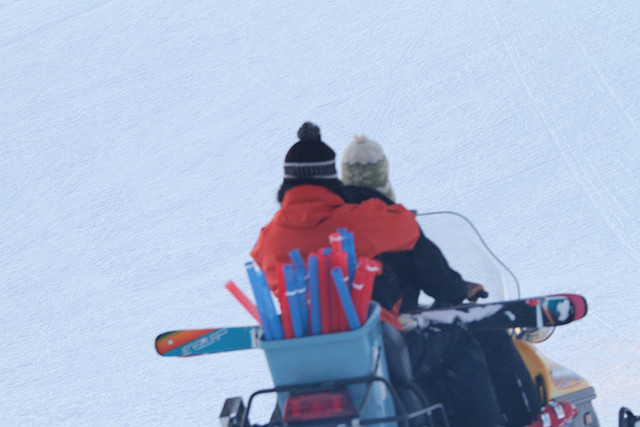}
& \hspace{-3mm}
\includegraphics[width=0.24\linewidth, height=30mm]{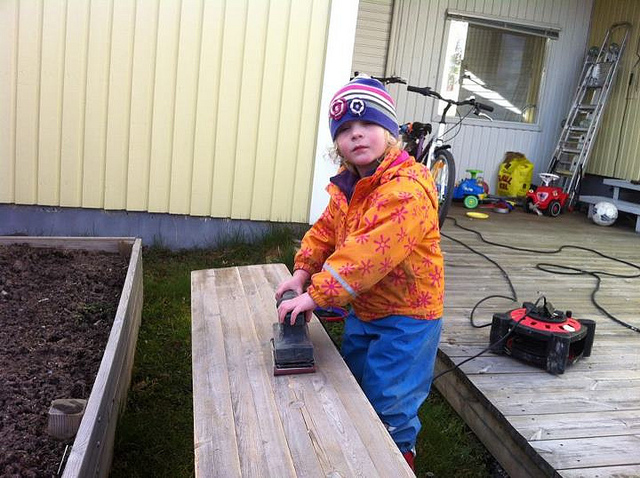}
& \hspace{-3mm}
\includegraphics[width=0.24\linewidth, height=30mm]{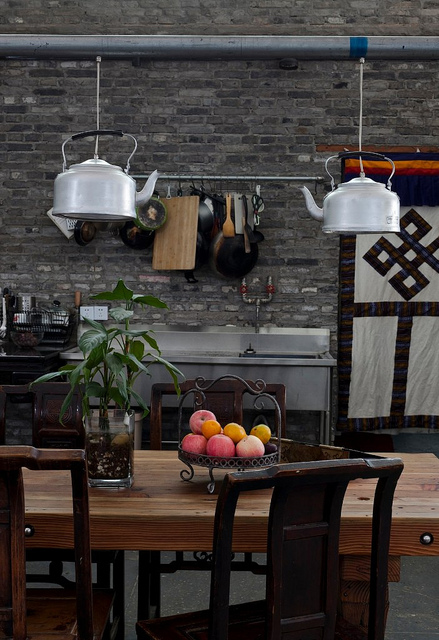}
& \hspace{-3mm}
\includegraphics[width=0.24\linewidth, height=30mm]{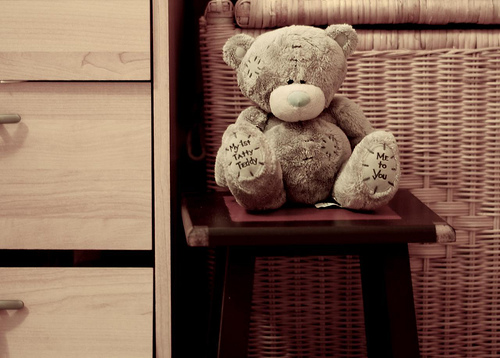}

\\
\vspace{-3mm}
\begin{tcolorbox}[width=0.24\linewidth, left=1pt,right=1pt,top=0pt,bottom=0pt]
how many people are in this photo? 
\end{tcolorbox}
& \hspace{-3mm}
\begin{tcolorbox}[width=0.24\linewidth, left=1pt,right=1pt,top=0pt,bottom=0pt]
what is leaning against the house?
\end{tcolorbox}
& \hspace{-3mm}
\begin{tcolorbox}[width=0.24\linewidth, left=1pt,right=1pt,top=0pt,bottom=0pt]
what has been upcycled to make lights?
\end{tcolorbox} 
& \hspace{-3mm}
\begin{tcolorbox}[width=0.24\linewidth, left=1pt,right=1pt,top=0pt,bottom=0pt]
what is the teddy bear sitting on? 
\end{tcolorbox}  \\

\vspace{-5mm}
\\
\begin{tcolorbox}[width=0.24\linewidth, left=1pt,right=1pt,top=0pt,bottom=0pt]
 TextMod \ityping: {\color{blue} 2} \\
 SpeechMod \ispeaking: {\color{blue} 2} 
\end{tcolorbox}
& \hspace{-3mm}
\begin{tcolorbox}[width=0.24\linewidth, left=1pt,right=1pt,top=0pt,bottom=0pt]
TextMod \ityping: {\color{red} tree} \\
SpeechMod \ispeaking: {\color{red} chair} 
\end{tcolorbox}
& \hspace{-3mm}
\begin{tcolorbox}[width=0.24\linewidth, left=1pt,right=1pt,top=0pt,bottom=0pt]
TextMod \ityping: {\color{red} bulb} \\
SpeechMod \ispeaking: {\color{red} bulb} 
\end{tcolorbox} 
& \hspace{-3mm}
\begin{tcolorbox}[width=0.24\linewidth, left=1pt,right=1pt,top=0pt,bottom=0pt]
TextMod \ityping: {\color{blue} chair} \\
SpeechMod \ispeaking: {\color{red} yes} 
\end{tcolorbox}  \\
\end{tabular}$
\vspace{-4mm}
\caption{Example of speech-based VQA answers. Correct answers in blue and incorrect in red.}  
\label{fig:visual:examples} 
\end{figure*}

A graphical version of the table is shown in \Cref{fig:noiseplots}. The constant values of SpeechMod \textit{Blind} and TextMod \textit{Blind} are included to show the noise level at which they perform better than their full model counterparts. Examples of the two models answering questions from the dataset are shown in \Cref{fig:visual:examples}.

One might imagine SpeechMod to perform better because of its direct optimization and end-to-end training solely for the task, yet this hypothesis does not hold true. At 0\% noise, TextMod achieves 7\% higher accuracy than SpeechMod. As noise is added, both models initially falter at similar rates, although their trends seem to head towards convergence. This is expected since, since at 100\% noise the question would not be audible at all; it would be random guessing, thus both methods would perform exactly the same.

\pgfplotsset{width=8cm,compat=1.9}
\begin{figure}[t]
\centering
\begin{tikzpicture}
\begin{axis}[
    title={},
    xlabel={Noise (\%)},
    ylabel={Accuracy (\%)},
    xmin=0, xmax=50,
    ymin=30, ymax=60,
    xtick={0,10,20,30,40,50},
    ytick={30,40,50,60},
    legend pos=north east,
    legend style={font=\fontsize{6}{6}\selectfont}
]
\addplot[
    color=blue,
    mark=square,
    ]
    coordinates {
    (0,46.99)(10,45.81)(20,43.33)(30,40.07)(40,35.85)(50,32.14)
    };
\addplot[
    color=magenta,
    mark=halfcircle
    ]
    coordinates {
    (0,54.03)(10,52.56)(20,50.22)(30,47.03)(40,42.83)(50,37.12)
    };
\addplot[mark=none, color=blue, dashed] coordinates{(0,42.05) (50,42.05)};
\addplot[mark=none, color=magenta, dashed] coordinates{(0,48.76) (50,48.76)};
\legend{SpeechMod, TextMod, SpeechMod Blind 0\% Noise, TextMod Blind \textit{OQ}}
\end{axis}
\end{tikzpicture}
\caption{SpeechMod and TextMod performance with varying amounts of added noise on \textit{test-dev}. \textit{Blind} counterparts are not tested on different noise levels.}
\label{fig:noiseplots}
\end{figure}
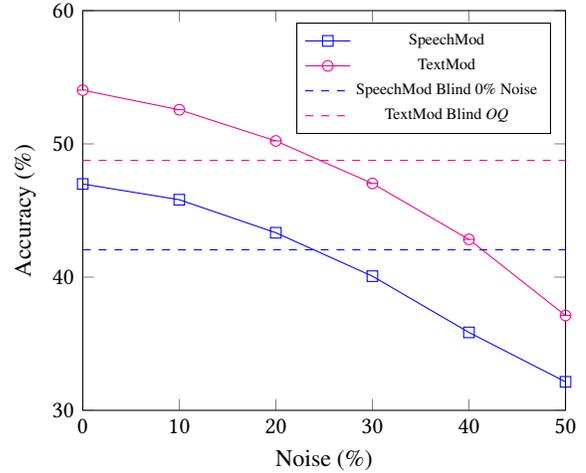

Next, we compare TextMod and SpeechMod against their respective \textit{Blind} models. The bias of questions in \textit{VQA1.0} is well documented. Namely, if the question is understood, there is a good chance of answering correctly without looking at the image (i.e. blind guessing). For example, \textit{Y/N} questions have the answer \emph{yes} more commonly than \emph{no}, so the system should guess \emph{yes} if a question is identified to be a \textit{Y/N} type. As a reference, always answering \emph{yes} yields a \textit{Y/N} accuracy of 70.81\% on \textit{test-dev}. The bias is clearly evident in both \textit{test-dev} and \textit{val} for TextMod and SpeechMod; the \textit{Y/N} section of \textit{Blind} always performs better than that of the 0\% data. Therefore, \textit{Blind} tells us how many questions are understood by these two modes of linguistic inputs. When comparing the linguistic only models with their complementary TextMod and SpeechMod, one can be certain that performances falling below the linguistic signifies that the model no longer understands the questions. Furthermore, perceiving the image and a noisy question becomes less informative than understanding a clean question without an image.

\subsection{Zero-Shot}
In this section zero-shot (ZS) results are analyzed to further understand the behavior of both models. ZS in the context of VQA refers to questions that were never seen in training. To get ZS data, we discard questions in \textit{val} subset that appeared in the \textit{train} subset, which decreased the number of valid questions from 104,654 to 65,365. Put differently, ZS is simply a subset of \textit{val}.

The models were trained on the complete \textit{train}, and best performing on the complete \textit{val} were selected. Next, the models were tested on the original and ZS datasets with noise injected (\Cref{table:zs}, \Cref{fig:noiseplots-zs}). These experiments were not be performed on \textit{test-dev} because the ground truth from \textit{test-dev} and \textit{test} are withheld and cannot be evaluated partially on the server.

As one would expect, ZS accuracies are worse than accuracies of the entire set, since models tend to perform more poorly on unseen data. TextMod performs 4\% better on the complete dataset than on the ZS, and SpeechMod on the complete dataset performs better by 7\%. The performance gap decreases as more noise is added. At 50\% noise, the performance on ZS and the original data have practically converged for both models. To the models, questions seen during the training but with high amount of noise added are as foreign as unseen questions.

\begin{table}[t]
\centering
\caption{Accuracy on \textit{zero-shot} with different levels of noise added. (Higher is better)}
\label{table:zs}
\begin{tabular}{ll|cccc}
          &                     & All    & Y/N    & Number & Other \\ \hline
TextMod   & OQ                  & 49.41  & 77.23  & 31.18  & 27.12 \\
          & 0\%                 & 46.41  & 73.37  & 30.64  & 24.38 \\
          & 10\%                & 45.23  & 71.93  & 30.32  & 23.24 \\
          & 20\%                & 43.30  & 69.26  & 29.63  & 21.75 \\
          & 30\%                & 40.85  & 65.84  & 28.55  & 19.89 \\
          & 40\%                & 37.79  & 62.56  & 26.10  & 16.91 \\
          & 50\%                & 34.41  & 59.58  & 21.50  & 13.42 \\ \hline
SpeechMod & 0\%                 & 37.01  & 65.58  & 23.19  & 12.99 \\
          & 10\%                & 36.52  & 65.12  & 22.83  & 12.45 \\
          & 20\%                & 35.47  & 64.04  & 22.29  & 11.29 \\
          & 30\%                & 34.08  & 62.77  & 21.45  &  9.67 \\
          & 40\%                & 32.12  & 60.59  & 19.94  &  7.81 \\
          & 50\%                & 29.88  & 57.70  & 18.20  &  6.09 
\end{tabular}
\end{table}

\subsection{Human Recordings}
Finally, a small, supplementary test is run on non-synthetic, human-recorded questions to see if the models would perform differently on real-world audio inputs. 1000 samples were randomly selected from \textit{val}, and the best performing models from the ZS section were used for evaluation. \Cref{table:recorded} shows the performance on the synthetic and human-recorded versions of this subset.

Although it is clear that both models have difficulties handling recorded questions, SpeechMod performs especially poorly. TextMod on the synthetic dataset achieves similar accuracy as it does on \textit{val} and \textit{test-dev} with 40\% noise. SpeechMod however, gets similar performance as the synthetic data with 50\% noise on only the \textit{Y/N} questions, while it seems to understand none of the other question types.

\subsection{Discussion}

As a modality, speech contains more information than text. In the process of reducing the high-dimensional audio inputs to the low-dimensional class output label (i.e. the answer), the best performing system must be that which extracts patterns most effectively.

TextMod relies heavily on the intermediate ASR system, which is more complicated than the entire architecture of SpeechMod, as the number of parameters one needs to learn for speech recognition is also much greater. The Kaldi model has also been trained on many times more data than contained in \textit{VQA1.0}. The ASR serves to filter out noise in high dimensions and extract meaningful patterns in the form of text. In a sense, one can think of the ASR as a feature extractor, with text being the salient feature and an explicit intermediate standardization of data before the question answering module.

Conversely, the only audio data SpeechMod learns from are the questions in the dataset. It does not include any mechanisms that explicitly learn semantics in a language, nor does it have intermediate data standardization. Thus, the model may not extract the concept of words from audio sounds. Whether or not forcing the system to learn words (i.e. transcribing words in the question and answering simultaneously) will be beneficial is left to future research, but it is evident that data standardization is helpful for unseen data.

\pgfplotsset{width=8cm,compat=1.9}
\begin{figure}[t]
\centering
\begin{tikzpicture}
\begin{axis}[
    title={},
    xlabel={Noise (\%)},
    ylabel={Accuracy (\%)},
    xmin=0, xmax=50,
    ymin=30, ymax=60,
    xtick={0,10,20,30,40,50},
    ytick={30,40,50,60},
    legend pos=north east,
    legend style={font=\fontsize{6}{6}\selectfont}
]

\addplot[
    color=blue,
    mark=none,
    dashed
    ]
    coordinates {
    ( 0,44.51)
    (10,43.62)
    (20,41.37)
    (30,37.95)
    (40,34.24)
    (50,30.54)
    };
\addplot[
    color=magenta,
    mark=none,
    dashed
    ]
    coordinates {
    ( 0,50.69)
    (10,49.28)
    (20,47.06)
    (30,43.99)
    (40,40.06)
    (50,35.54)
    };
\addplot[
    color=blue,
    mark=square
    ]
    coordinates {
    ( 0,37.01)
    (10,36.52)
    (20,35.47)
    (30,34.08)
    (40,32.12)
    (50,29.88)
    };
\addplot[
    color=magenta,
    mark=halfcircle
    ]
    coordinates {
    ( 0,46.41)
    (10,45.23)
    (20,43.30)
    (30,40.85)
    (40,37.79)
    (50,34.41)
    };
\legend{SpeechMod \textit{val}, TextMod \textit{val}, SpeechMod ZS, TextMod ZS}
\end{axis}
\end{tikzpicture}
\caption{SpeechMod and TextMod performance with varying amounts of added noise for \textit{zero-shot} subset in reference to the complete datasets.}
\label{fig:noiseplots-zs}
\end{figure}
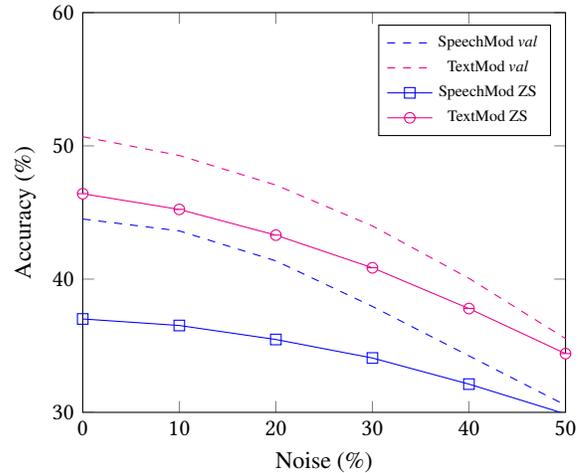

\begin{table}[t]
\centering
\caption{Performance on 1000 human-recorded questions.}
\label{table:recorded}
\begin{tabular}{l|cccc}
                        & All   & Y/N   & Number & Other \\ \hline
SpeechMod (Recorded)    & 21.46 & 57.26 & 0.77   & 0.91  \\
SpeechMod (Synthetic)   & 42.69 & 66.58 & 32.31  & 27.58 \\ \hline
TextMod (Recorded)      & 41.66 & 66.33 & 35.69  & 25.37 \\
TextMod (Original Text) & 53.09 & 77.73 & 41.54  & 38.26
\end{tabular}
\end{table}

In ZS experiments, the gap in performance between the unseen and full dataset with TextMod is much smaller than in SpeechMod (4\% vs 7\%). A text-based system can still glimpse the meaning of a question even if a word has never been seen, but from the perspective of SpeechMod, new words represent entirely different signal trajectories. Furthermore, audio inputs are continuous streams, making it difficult to differentiate when the new words begin or end. 

A similar effect is amplified in answering human-recorded questions. The synthetic audio sounds monotonous, disinterested, with little silence between words while the human-recorded audio has inflections, emphasis, accents, and pauses. An inspection of the spectrograms (\Cref{fig:specs}) confirms this, as the synthetic waveforms have vastly different audio signatures. Because SpeechMod has no training data similar to the human-recorded samples, it is unable to extract salient patterns. In comparison, the ASR removes most of the variance in the input by standardizing the audio into a compact, salient textual representation. From the perspective of TextMod, the human-recorded questions is only slightly different than those provided in training.

It is evident in our experiments that text-based VQA performs better than speech-based, but bearing in mind the simple architecture and limited amount of training data, we believe the results of SpeechMod merits further study into end-to-end methods.

\subsection{Future Work}
As alluded to in previous sections, there are a few research directions that may yield interesting results. One straightforward approach to improving the end-to-end model is by data augmentation. It is widely accepted that effectiveness of neural architectures is data driven, so training with noisy data and different speakers will make the model more robust to inputs during run time. Just as many possibilities exist in improving the architecture. One can add feature extractors, attention mechanisms, GAN training, or any amalgamation of the techniques in the deep learning mainstream. An interesting study would be to enforce the prediction of the question while simultaneously learning to answer the question. Doing so may improve performance, but more importantly allows us to interpret the concepts learned by the neural network.

Another direction is to restrict the amount of training data available to both approaches to observe their learning efficiency. For example, minor languages may not have a reliable ASR. One can simulate a minor language by training an ASR with only the data available in the training set, and comparing this approach with the end-to-end method trained on the same amount of data.

\section{Conclusion}

We have proposed speech-based visual question answering and introduced two approaches that tackle this problem, one of which can be trained end-to-end on audio inputs. Despite its simple architecture, the end-to-end method works well when the test data has audio signatures comparable to its training data. Both methods suffered performance decreases at similar rates when noise is introduced. A pipelined method using an ASR tolerates varied inputs much better because it normalizes the input variance into text before running the VQA module. We release the speech dataset and invite the multimedia research community to explore the intersection of speech and vision.

\bibliographystyle{ACM-Reference-Format}
\bibliography{sigproc} 


\begin{thebibliography}{00}


\ifx \showCODEN    \undefined \def \showCODEN     #1{\unskip}     \fi
\ifx \showDOI      \undefined \def \showDOI       #1{{\tt DOI:}\penalty0{#1}\ }
  \fi
\ifx \showISBNx    \undefined \def \showISBNx     #1{\unskip}     \fi
\ifx \showISBNxiii \undefined \def \showISBNxiii  #1{\unskip}     \fi
\ifx \showISSN     \undefined \def \showISSN      #1{\unskip}     \fi
\ifx \showLCCN     \undefined \def \showLCCN      #1{\unskip}     \fi
\ifx \shownote     \undefined \def \shownote      #1{#1}          \fi
\ifx \showarticletitle \undefined \def \showarticletitle #1{#1}   \fi
\ifx \showURL      \undefined \def \showURL       {\relax}        \fi
\providecommand\bibfield[2]{#2}
\providecommand\bibinfo[2]{#2}
\providecommand\natexlab[1]{#1}
\providecommand\showeprint[2][]{arXiv:#2}

\bibitem[\protect\citeauthoryear{??}{smi}{2015}]%
        {smile}
 \bibinfo{year}{2015}\natexlab{}.
\newblock \bibinfo{title}{Smile - Smart Photo Annotation}.
\newblock
  \bibinfo{howpublished}{\url{https://play.google.com/store/apps/details?id=com.neuromorphic.retinet.smile}}.
    (\bibinfo{year}{2015}).
\newblock


\bibitem[\protect\citeauthoryear{Amodei, Anubhai, Battenberg, Case, Casper,
  Catanzaro, Chen, Chrzanowski, Coates, Diamos, et~al\mbox{.}}{Amodei
  et~al\mbox{.}}{2015}]%
        {deepspeech2}
\bibfield{author}{\bibinfo{person}{Dario Amodei}, \bibinfo{person}{Rishita
  Anubhai}, \bibinfo{person}{Eric Battenberg}, \bibinfo{person}{Carl Case},
  \bibinfo{person}{Jared Casper}, \bibinfo{person}{Bryan Catanzaro},
  \bibinfo{person}{Jingdong Chen}, \bibinfo{person}{Mike Chrzanowski},
  \bibinfo{person}{Adam Coates}, \bibinfo{person}{Greg Diamos}, {and}
  \bibinfo{person}{others}.} \bibinfo{year}{2015}\natexlab{}.
\newblock \showarticletitle{Deep speech 2: End-to-end speech recognition in
  english and mandarin}.
\newblock \bibinfo{journal}{{\em arXiv preprint arXiv:1512.02595\/}}
  (\bibinfo{year}{2015}).
\newblock


\bibitem[\protect\citeauthoryear{Antol, Agrawal, Lu, Mitchell, Batra, Zitnick,
  and Parikh}{Antol et~al\mbox{.}}{2015}]%
        {VQA}
\bibfield{author}{\bibinfo{person}{Stanislaw Antol}, \bibinfo{person}{Aishwarya
  Agrawal}, \bibinfo{person}{Jiasen Lu}, \bibinfo{person}{Margaret Mitchell},
  \bibinfo{person}{Dhruv Batra}, \bibinfo{person}{C.~Lawrence Zitnick}, {and}
  \bibinfo{person}{Devi Parikh}.} \bibinfo{year}{2015}\natexlab{}.
\newblock \showarticletitle{VQA: Visual Question Answering}. In
  \bibinfo{booktitle}{{\em International Conference on Computer Vision
  (ICCV)}}.
\newblock


\bibitem[\protect\citeauthoryear{Aytar, Vondrick, and Torralba}{Aytar
  et~al\mbox{.}}{2016}]%
        {soundnet}
\bibfield{author}{\bibinfo{person}{Yusuf Aytar}, \bibinfo{person}{Carl
  Vondrick}, {and} \bibinfo{person}{Antonio Torralba}.}
  \bibinfo{year}{2016}\natexlab{}.
\newblock \showarticletitle{SoundNet: Learning Sound Representations from
  Unlabeled Video}.
\newblock \bibinfo{journal}{{\em CoRR\/}}  \bibinfo{volume}{abs/1610.09001}
  (\bibinfo{year}{2016}).
\newblock
\showURL{%
\url{http://arxiv.org/abs/1610.09001}}


\bibitem[\protect\citeauthoryear{Bahdanau, Cho, and Bengio}{Bahdanau
  et~al\mbox{.}}{2014}]%
        {nmt:joint:align}
\bibfield{author}{\bibinfo{person}{Dzmitry Bahdanau},
  \bibinfo{person}{Kyunghyun Cho}, {and} \bibinfo{person}{Yoshua Bengio}.}
  \bibinfo{year}{2014}\natexlab{}.
\newblock \showarticletitle{Neural machine translation by jointly learning to
  align and translate}.
\newblock \bibinfo{journal}{{\em arXiv preprint arXiv:1409.0473\/}}
  (\bibinfo{year}{2014}).
\newblock


\bibitem[\protect\citeauthoryear{Bigham, Jayant, Ji, Little, Miller, Miller,
  Miller, Tatarowicz, White, White, et~al\mbox{.}}{Bigham
  et~al\mbox{.}}{2010}]%
        {realtime:vqa}
\bibfield{author}{\bibinfo{person}{Jeffrey~P Bigham},
  \bibinfo{person}{Chandrika Jayant}, \bibinfo{person}{Hanjie Ji},
  \bibinfo{person}{Greg Little}, \bibinfo{person}{Andrew Miller},
  \bibinfo{person}{Robert~C Miller}, \bibinfo{person}{Robin Miller},
  \bibinfo{person}{Aubrey Tatarowicz}, \bibinfo{person}{Brandyn White},
  \bibinfo{person}{Samual White}, {and} \bibinfo{person}{others}.}
  \bibinfo{year}{2010}\natexlab{}.
\newblock \showarticletitle{VizWiz: nearly real-time answers to visual
  questions}. In \bibinfo{booktitle}{{\em Proceedings of the 23nd annual ACM
  symposium on User interface software and technology}}. ACM,
  \bibinfo{pages}{333--342}.
\newblock


\bibitem[\protect\citeauthoryear{Cheng, Zheng, Lin, Vineet, Sturgess, Crook,
  Mitra, and Torr}{Cheng et~al\mbox{.}}{2014}]%
        {image:spirit}
\bibfield{author}{\bibinfo{person}{Ming-Ming Cheng}, \bibinfo{person}{Shuai
  Zheng}, \bibinfo{person}{Wen-Yan Lin}, \bibinfo{person}{Vibhav Vineet},
  \bibinfo{person}{Paul Sturgess}, \bibinfo{person}{Nigel Crook},
  \bibinfo{person}{Niloy~J. Mitra}, {and} \bibinfo{person}{Philip Torr}.}
  \bibinfo{year}{2014}\natexlab{}.
\newblock \showarticletitle{ImageSpirit: Verbal Guided Image Parsing}.
\newblock \bibinfo{journal}{{\em ACM Trans. Graph.\/}} \bibinfo{volume}{34},
  \bibinfo{number}{1} (\bibinfo{year}{2014}), \bibinfo{pages}{3:1--3:11}.
\newblock


\bibitem[\protect\citeauthoryear{Fern{\'a}ndez, Graves, and
  Schmidhuber}{Fern{\'a}ndez et~al\mbox{.}}{2007}]%
        {rnn:discrimspotting}
\bibfield{author}{\bibinfo{person}{Santiago Fern{\'a}ndez},
  \bibinfo{person}{Alex Graves}, {and} \bibinfo{person}{J{\"u}rgen
  Schmidhuber}.} \bibinfo{year}{2007}\natexlab{}.
\newblock \bibinfo{booktitle}{{\em An Application of Recurrent Neural Networks
  to Discriminative Keyword Spotting}}.
\newblock \bibinfo{publisher}{Springer Berlin Heidelberg},
  \bibinfo{address}{Berlin, Heidelberg}, \bibinfo{pages}{220--229}.
\newblock
\showISBNx{978-3-540-74695-9}
\showDOI{%
\url{https://doi.org/10.1007/978-3-540-74695-9_23}}


\bibitem[\protect\citeauthoryear{Fukui, Park, Yang, Rohrbach, Darrell, and
  Rohrbach}{Fukui et~al\mbox{.}}{2016}]%
        {multimodal:pooling}
\bibfield{author}{\bibinfo{person}{Akira Fukui}, \bibinfo{person}{Dong~Huk
  Park}, \bibinfo{person}{Daylen Yang}, \bibinfo{person}{Anna Rohrbach},
  \bibinfo{person}{Trevor Darrell}, {and} \bibinfo{person}{Marcus Rohrbach}.}
  \bibinfo{year}{2016}\natexlab{}.
\newblock \showarticletitle{Multimodal compact bilinear pooling for visual
  question answering and visual grounding}.
\newblock \bibinfo{journal}{{\em arXiv preprint arXiv:1606.01847\/}}
  (\bibinfo{year}{2016}).
\newblock


\bibitem[\protect\citeauthoryear{Graves, Fern{\'a}ndez, Gomez, and
  Schmidhuber}{Graves et~al\mbox{.}}{2006}]%
        {ctc}
\bibfield{author}{\bibinfo{person}{Alex Graves}, \bibinfo{person}{Santiago
  Fern{\'a}ndez}, \bibinfo{person}{Faustino Gomez}, {and}
  \bibinfo{person}{J{\"u}rgen Schmidhuber}.} \bibinfo{year}{2006}\natexlab{}.
\newblock \showarticletitle{Connectionist temporal classification: labelling
  unsegmented sequence data with recurrent neural networks}. In
  \bibinfo{booktitle}{{\em Proceedings of the 23rd international conference on
  Machine learning}}. ACM, \bibinfo{pages}{369--376}.
\newblock


\bibitem[\protect\citeauthoryear{Harwath, Torralba, and Glass}{Harwath
  et~al\mbox{.}}{2016}]%
        {speech:caption}
\bibfield{author}{\bibinfo{person}{David Harwath}, \bibinfo{person}{Antonio
  Torralba}, {and} \bibinfo{person}{James Glass}.}
  \bibinfo{year}{2016}\natexlab{}.
\newblock \showarticletitle{Unsupervised Learning of Spoken Language with
  Visual Context}. In \bibinfo{booktitle}{{\em Advances in Neural Information
  Processing Systems}}. \bibinfo{pages}{1858--1866}.
\newblock


\bibitem[\protect\citeauthoryear{Hazen, Sherry, and Adler}{Hazen
  et~al\mbox{.}}{2007}]%
        {speech:retri:img}
\bibfield{author}{\bibinfo{person}{Timothy~J. Hazen}, \bibinfo{person}{Brennan
  Sherry}, {and} \bibinfo{person}{Mark Adler}.}
  \bibinfo{year}{2007}\natexlab{}.
\newblock \showarticletitle{Speech-based annotation and retrieval of digital
  photographs}. In \bibinfo{booktitle}{{\em INTERSPEECH}}.
\newblock


\bibitem[\protect\citeauthoryear{Kalashnikov, Mehrotra, Xu, and
  Venkatasubramanian}{Kalashnikov et~al\mbox{.}}{2011}]%
        {speech:anno:img}
\bibfield{author}{\bibinfo{person}{D.V. Kalashnikov}, \bibinfo{person}{S.
  Mehrotra}, \bibinfo{person}{Jie Xu}, {and} \bibinfo{person}{N.
  Venkatasubramanian}.} \bibinfo{year}{2011}\natexlab{}.
\newblock \showarticletitle{A Semantics-Based Approach for Speech Annotation of
  Images}.
\newblock \bibinfo{journal}{{\em IEEE Trans. Knowl. Data Eng.\/}}
  \bibinfo{volume}{23}, \bibinfo{number}{9} (\bibinfo{year}{2011}),
  \bibinfo{pages}{1373--1387}.
\newblock


\bibitem[\protect\citeauthoryear{Kingma and Ba}{Kingma and Ba}{2014}]%
        {kingma2014adam}
\bibfield{author}{\bibinfo{person}{Diederik Kingma} {and}
  \bibinfo{person}{Jimmy Ba}.} \bibinfo{year}{2014}\natexlab{}.
\newblock \showarticletitle{Adam: A method for stochastic optimization}.
\newblock \bibinfo{journal}{{\em arXiv preprint arXiv:1412.6980\/}}
  (\bibinfo{year}{2014}).
\newblock


\bibitem[\protect\citeauthoryear{Laput, Dontcheva, Wilensky, Chang, Agarwala,
  Linder, and Adar}{Laput et~al\mbox{.}}{2013}]%
        {pixel:tone}
\bibfield{author}{\bibinfo{person}{Gierad~P Laput}, \bibinfo{person}{Mira
  Dontcheva}, \bibinfo{person}{Gregg Wilensky}, \bibinfo{person}{Walter Chang},
  \bibinfo{person}{Aseem Agarwala}, \bibinfo{person}{Jason Linder}, {and}
  \bibinfo{person}{Eytan Adar}.} \bibinfo{year}{2013}\natexlab{}.
\newblock \showarticletitle{PixelTone: a multimodal interface for image
  editing}. In \bibinfo{booktitle}{{\em CHI}}.
\newblock


\bibitem[\protect\citeauthoryear{LeCun, Bengio, and Hinton}{LeCun
  et~al\mbox{.}}{2015}]%
        {lecun2015deep}
\bibfield{author}{\bibinfo{person}{Yann LeCun}, \bibinfo{person}{Yoshua
  Bengio}, {and} \bibinfo{person}{Geoffrey Hinton}.}
  \bibinfo{year}{2015}\natexlab{}.
\newblock \showarticletitle{Deep learning}.
\newblock \bibinfo{journal}{{\em Nature\/}} \bibinfo{volume}{521},
  \bibinfo{number}{7553} (\bibinfo{year}{2015}), \bibinfo{pages}{436--444}.
\newblock


\bibitem[\protect\citeauthoryear{Lu, Yang, Batra, and Parikh}{Lu
  et~al\mbox{.}}{2016}]%
        {vqa:hieco}
\bibfield{author}{\bibinfo{person}{Jiasen Lu}, \bibinfo{person}{Jianwei Yang},
  \bibinfo{person}{Dhruv Batra}, {and} \bibinfo{person}{Devi Parikh}.}
  \bibinfo{year}{2016}\natexlab{}.
\newblock \bibinfo{title}{Hierarchical Question-Image Co-Attention for Visual
  Question Answering}.
\newblock   (\bibinfo{year}{2016}).
\newblock


\bibitem[\protect\citeauthoryear{Malinowski and Fritz}{Malinowski and
  Fritz}{2014}]%
        {daquar}
\bibfield{author}{\bibinfo{person}{Mateusz Malinowski} {and}
  \bibinfo{person}{Mario Fritz}.} \bibinfo{year}{2014}\natexlab{}.
\newblock \showarticletitle{A multi-world approach to question answering about
  real-world scenes based on uncertain input}. In \bibinfo{booktitle}{{\em
  Advances in Neural Information Processing Systems}}.
  \bibinfo{pages}{1682--1690}.
\newblock


\bibitem[\protect\citeauthoryear{Mnih, Heess, Graves, et~al\mbox{.}}{Mnih
  et~al\mbox{.}}{2014}]%
        {recurrent:visual:attn}
\bibfield{author}{\bibinfo{person}{Volodymyr Mnih}, \bibinfo{person}{Nicolas
  Heess}, \bibinfo{person}{Alex Graves}, {and} \bibinfo{person}{others}.}
  \bibinfo{year}{2014}\natexlab{}.
\newblock \showarticletitle{Recurrent models of visual attention}. In
  \bibinfo{booktitle}{{\em Advances in neural information processing systems}}.
  \bibinfo{pages}{2204--2212}.
\newblock


\bibitem[\protect\citeauthoryear{Nam, Ha, and Kim}{Nam et~al\mbox{.}}{2016}]%
        {vqa:dualattn}
\bibfield{author}{\bibinfo{person}{Hyeonseob Nam}, \bibinfo{person}{Jung-Woo
  Ha}, {and} \bibinfo{person}{Jeonghee Kim}.} \bibinfo{year}{2016}\natexlab{}.
\newblock \showarticletitle{Dual Attention Networks for Multimodal Reasoning
  and Matching}.
\newblock \bibinfo{journal}{{\em CoRR\/}}  \bibinfo{volume}{abs/1611.00471}
  (\bibinfo{year}{2016}).
\newblock


\bibitem[\protect\citeauthoryear{Povey, Ghoshal, Boulianne, Burget, Glembek,
  Goel, Hannemann, Motlicek, Qian, Schwarz, Silovsky, Stemmer, and
  Vesely}{Povey et~al\mbox{.}}{2011}]%
        {kaldi}
\bibfield{author}{\bibinfo{person}{Daniel Povey}, \bibinfo{person}{Arnab
  Ghoshal}, \bibinfo{person}{Gilles Boulianne}, \bibinfo{person}{Lukas Burget},
  \bibinfo{person}{Ondrej Glembek}, \bibinfo{person}{Nagendra Goel},
  \bibinfo{person}{Mirko Hannemann}, \bibinfo{person}{Petr Motlicek},
  \bibinfo{person}{Yanmin Qian}, \bibinfo{person}{Petr Schwarz},
  \bibinfo{person}{Jan Silovsky}, \bibinfo{person}{Georg Stemmer}, {and}
  \bibinfo{person}{Karel Vesely}.} \bibinfo{year}{2011}\natexlab{}.
\newblock \showarticletitle{The Kaldi Speech Recognition Toolkit}. In
  \bibinfo{booktitle}{{\em IEEE 2011 Workshop on Automatic Speech Recognition
  and Understanding}}. \bibinfo{publisher}{IEEE Signal Processing Society}.
\newblock
\newblock
\shownote{IEEE Catalog No.: CFP11SRW-USB.}


\bibitem[\protect\citeauthoryear{Ruan, Wobbrock, Liou, Ng, and Landay}{Ruan
  et~al\mbox{.}}{2016}]%
        {speech:faster}
\bibfield{author}{\bibinfo{person}{Sherry Ruan}, \bibinfo{person}{Jacob~O
  Wobbrock}, \bibinfo{person}{Kenny Liou}, \bibinfo{person}{Andrew Ng}, {and}
  \bibinfo{person}{James Landay}.} \bibinfo{year}{2016}\natexlab{}.
\newblock \showarticletitle{Speech Is 3x Faster than Typing for English and
  Mandarin Text Entry on Mobile Devices}.
\newblock \bibinfo{journal}{{\em arXiv preprint arXiv:1608.07323\/}}
  (\bibinfo{year}{2016}).
\newblock


\bibitem[\protect\citeauthoryear{Russakovsky, Deng, Su, Krause, Satheesh, Ma,
  Huang, Karpathy, Khosla, Bernstein, et~al\mbox{.}}{Russakovsky
  et~al\mbox{.}}{2015}]%
        {imagenet:2015}
\bibfield{author}{\bibinfo{person}{Olga Russakovsky}, \bibinfo{person}{Jia
  Deng}, \bibinfo{person}{Hao Su}, \bibinfo{person}{Jonathan Krause},
  \bibinfo{person}{Sanjeev Satheesh}, \bibinfo{person}{Sean Ma},
  \bibinfo{person}{Zhiheng Huang}, \bibinfo{person}{Andrej Karpathy},
  \bibinfo{person}{Aditya Khosla}, \bibinfo{person}{Michael Bernstein}, {and}
  \bibinfo{person}{others}.} \bibinfo{year}{2015}\natexlab{}.
\newblock \showarticletitle{Imagenet large scale visual recognition challenge}.
\newblock \bibinfo{journal}{{\em International Journal of Computer Vision\/}}
  \bibinfo{volume}{115}, \bibinfo{number}{3} (\bibinfo{year}{2015}),
  \bibinfo{pages}{211--252}.
\newblock


\bibitem[\protect\citeauthoryear{Salamon, Jacoby, and Bello}{Salamon
  et~al\mbox{.}}{2014}]%
        {urban8k}
\bibfield{author}{\bibinfo{person}{Justin Salamon},
  \bibinfo{person}{Christopher Jacoby}, {and} \bibinfo{person}{Juan~Pablo
  Bello}.} \bibinfo{year}{2014}\natexlab{}.
\newblock \showarticletitle{A Dataset and Taxonomy for Urban Sound Research}.
  In \bibinfo{booktitle}{{\em Proceedings of the {ACM} International Conference
  on Multimedia, {MM} '14, Orlando, FL, USA, November 03 - 07, 2014}}.
  \bibinfo{pages}{1041--1044}.
\newblock
\showDOI{%
\url{https://doi.org/10.1145/2647868.2655045}}


\bibitem[\protect\citeauthoryear{Simonyan and Zisserman}{Simonyan and
  Zisserman}{2015}]%
        {vgg16}
\bibfield{author}{\bibinfo{person}{K. Simonyan} {and} \bibinfo{person}{A.
  Zisserman}.} \bibinfo{year}{2015}\natexlab{}.
\newblock \showarticletitle{Very Deep Convolutional Networks for Large-Scale
  Image Recognition}. In \bibinfo{booktitle}{{\em ICLR}}.
\newblock


\bibitem[\protect\citeauthoryear{Srihari and Zhang}{Srihari and Zhang}{2000}]%
        {show:tell}
\bibfield{author}{\bibinfo{person}{R.K. Srihari} {and}
  \bibinfo{person}{Zhongfei Zhang}.} \bibinfo{year}{2000}\natexlab{}.
\newblock \showarticletitle{Show\&Tell: a semi-automated image annotation
  system}.
\newblock \bibinfo{journal}{{\em MultiMedia, IEEE\/}} \bibinfo{volume}{7},
  \bibinfo{number}{3} (\bibinfo{year}{2000}), \bibinfo{pages}{61--71}.
\newblock


\bibitem[\protect\citeauthoryear{Vinyals, Toshev, Bengio, and Erhan}{Vinyals
  et~al\mbox{.}}{2015}]%
        {show:tell:caption}
\bibfield{author}{\bibinfo{person}{Oriol Vinyals}, \bibinfo{person}{Alexander
  Toshev}, \bibinfo{person}{Samy Bengio}, {and} \bibinfo{person}{Dumitru
  Erhan}.} \bibinfo{year}{2015}\natexlab{}.
\newblock \showarticletitle{Show and Tell: A Neural Image Caption Generator}.
  In \bibinfo{booktitle}{{\em CVPR}}.
\newblock


\bibitem[\protect\citeauthoryear{Xiong, Merity, and Socher}{Xiong
  et~al\mbox{.}}{2016}]%
        {vqa:dynamicmemory}
\bibfield{author}{\bibinfo{person}{Caiming Xiong}, \bibinfo{person}{Stephen
  Merity}, {and} \bibinfo{person}{Richard Socher}.}
  \bibinfo{year}{2016}\natexlab{}.
\newblock \showarticletitle{Dynamic memory networks for visual and textual
  question answering}.
\newblock \bibinfo{journal}{{\em arXiv\/}}  \bibinfo{volume}{1603}
  (\bibinfo{year}{2016}).
\newblock


\bibitem[\protect\citeauthoryear{Xu and Saenko}{Xu and Saenko}{2016}]%
        {vqa:spatial:attn}
\bibfield{author}{\bibinfo{person}{Huijuan Xu} {and} \bibinfo{person}{Kate
  Saenko}.} \bibinfo{year}{2016}\natexlab{}.
\newblock \showarticletitle{Ask, attend and answer: Exploring question-guided
  spatial attention for visual question answering}. In \bibinfo{booktitle}{{\em
  European Conference on Computer Vision}}. Springer,
  \bibinfo{pages}{451--466}.
\newblock


\bibitem[\protect\citeauthoryear{Yang, He, Gao, Deng, and Smola}{Yang
  et~al\mbox{.}}{2016}]%
        {vqa:stackedattn}
\bibfield{author}{\bibinfo{person}{Zichao Yang}, \bibinfo{person}{Xiaodong He},
  \bibinfo{person}{Jianfeng Gao}, \bibinfo{person}{Li Deng}, {and}
  \bibinfo{person}{Alex Smola}.} \bibinfo{year}{2016}\natexlab{}.
\newblock \showarticletitle{Stacked attention networks for image question
  answering}. In \bibinfo{booktitle}{{\em Proceedings of the IEEE Conference on
  Computer Vision and Pattern Recognition}}. \bibinfo{pages}{21--29}.
\newblock


\bibitem[\protect\citeauthoryear{Zhu, Kiros, Zemel, Salakhutdinov, Urtasun,
  Torralba, and Fidler}{Zhu et~al\mbox{.}}{2015}]%
        {align:bookmovie}
\bibfield{author}{\bibinfo{person}{Yukun Zhu}, \bibinfo{person}{Ryan Kiros},
  \bibinfo{person}{Rich Zemel}, \bibinfo{person}{Ruslan Salakhutdinov},
  \bibinfo{person}{Raquel Urtasun}, \bibinfo{person}{Antonio Torralba}, {and}
  \bibinfo{person}{Sanja Fidler}.} \bibinfo{year}{2015}\natexlab{}.
\newblock \showarticletitle{Aligning books and movies: Towards story-like
  visual explanations by watching movies and reading books}. In
  \bibinfo{booktitle}{{\em Proceedings of the IEEE International Conference on
  Computer Vision}}. \bibinfo{pages}{19--27}.
\newblock


\end{thebibliography}

\end{document}